\documentclass{article}
\usepackage{spconf,amsmath,graphicx}
\usepackage{booktabs}
\usepackage{makecell}
\usepackage[inkscapeformat=png]{svg}
\usepackage{subcaption}
\usepackage{cleveref}

\usepackage{enumitem}
\setlist{nosep, leftmargin=14pt}

\usepackage{mwe} % to get dummy images

\usepackage{pifont}% http://ctan.org/pkg/pifont
\newcommand{\cmark}{\ding{51}}%
\newcommand{\xmark}{\ding{55}}%

\name{
\begin{minipage}{\textwidth}
\centering
Lena Heinemann\textsuperscript{1*} \qquad Alexander Jaus\textsuperscript{1*}\thanks{$^{\star}$ indicates equal paper contribution} \qquad Zdravko Marinov\textsuperscript{1}  \qquad \\
 Moon Kim\textsuperscript{2}  \qquad Maria Francesca Spadea\textsuperscript{3} \qquad Jens Kleesiek\textsuperscript{2} \qquad Rainer Stiefelhagen\textsuperscript{1}
\end{minipage}
}

\address{\textsuperscript{1} Institute for Anthropomatics \& Robotics (IAR), Karlsruhe Institute of Technology, Germany \\
         \textsuperscript{2} Institute for AI in Medicine (IKIM), University Hospital Essen, Germany \\
         \textsuperscript{3} Insitute of Biomedical Engineering (IBT), Karlsruhe Insitute of Technology, Germany
         }

\title{LIMIS: Towards Language-based Interactive Medical Image Segmentation}
\begin{document}

\maketitle

\begin{abstract}
%100 to 150 word
Within this work, we introduce LIMIS: The first purely language-based interactive medical image segmentation model. We achieve this by adapting Grounded SAM to the medical domain and designing a language-based model interaction strategy that allows radiologists to incorporate their knowledge into the segmentation process. LIMIS produces high-quality initial segmentation masks by leveraging medical foundation models and allows users to adapt segmentation masks using only language, opening up interactive segmentation to scenarios where physicians require using their hands for other tasks. We evaluate LIMIS on three publicly available medical datasets in terms of performance and usability with experts from the medical domain confirming its high-quality segmentation masks and its interactive usability. 

%Medical image segmentation is a crucial step in the processing of medical images. If it is done by hand, it is labor-intensive and time-consuming. We present a novel segmentation pipeline that enables interactive medical image segmentation using only text prompts. Our approach adapts Grounding DINO to the medical domain and combines it with SAM to Grounded SAM. Users can iteratively refine segmentation masks through a series of text-based interactions, allowing them to incorporate their knowledge without the need for physical interaction. This enables users to create segmentation masks in situations when they can not use physical interactions. We evaluate our pipeline on \textbf{X} public datasets and demonstrate the usability and potential of the approach with a user study.
\end{abstract}
\section{Introduction and Related Work}
\label{sec:intro}
Semantic Segmentation has become an essential tool in many automated clinical applications. It enriches medical images with pixel-wise semantic meaning allowing downstream applications and physicians to assess the precise location and type of anatomical structures or pathological regions. The image segmentation process is, however, if done by hand a labor-intensive and time-consuming process. While neural network-based segmentations offer some form of speedup by generating automated segmentation masks, initial results are often unsatisfying due to insufficient quality, noisy data, or unexpected distribution shifts. Even when assuming a perfect prediction, a network may be trained under a different annotation protocol (e.g. liver vessels are treated as part of the liver instead of a separate class) which may be undesired for the current application. Interactive segmentation which puts the physician in the loop with a network can mitigate the described problems. It combines user interactions with automatic algorithms allowing physicians to contribute their expert knowledge. %The interactions can be bounding boxes \cite{rajchl_deepcut_2016}, clicks \cite{shahedi_technical_2022}, scribbles \cite{wong_scribbleprompt_2023} or text \cite{wu_phrasecut_nodate}, for example. An initially generated segmentation is often not sufficient. This may be due to the inaccurate segmentation of the automatic algorithm or because the user has individual demands on the segmentation mask. 

\begin{table}[t]

\setlength{\tabcolsep}{4.2pt}
\begin{minipage}[b]{0.65\linewidth}
  \centering
    \begin{tabular}{lcccc}
   
    % \toprule
    Model & \rotatebox{90}{Interactive} & \rotatebox{90}{Physical Interact.} & \rotatebox{90}{Lang.-based Seg.} &  \rotatebox{90}{Lang.-based Interact.} \\
    \midrule
    nnUNet~\cite{isensee_nnu-net_2021} & \xmark & - & - & - \\
    TotalSeg.~\cite{wasserthal_totalsegmentator_2022} & \xmark & - & - & - \\
    SAM-based~\cite{kirillov_segment_2023} & \cmark & \cmark & \xmark & \xmark \\
    ScribblePromt~\cite{wong_scribbleprompt_2023} & \cmark & \cmark & \xmark & \xmark \\
    GroundedSAM~\cite{idea_groundedsam_2024} & \cmark & \xmark & \cmark & \xmark \\

    \midrule
    LIMIS (ours) & \cmark & \xmark & \cmark & \cmark \\
    \bottomrule
    \\
    \end{tabular}
\medskip
\end{minipage}
\begin{minipage}[t]{0.34\linewidth}
  \centering
    \includegraphics[width=\linewidth, trim=0 260 830 0, clip]{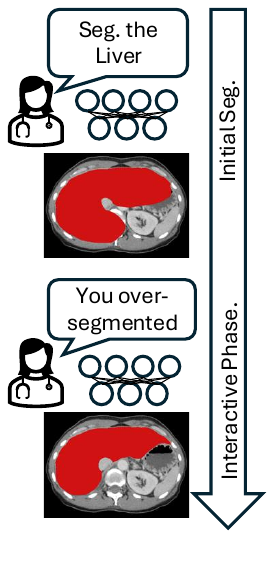}

\end{minipage}
%\end{center}
\vspace{-0.7cm}
\caption{LIMIS offers a unique and purely natural-language-based segmentation and interaction strategy.}
\vspace{-0.3cm}
 \label{fig:title-fig}
\end{table}
%\vspace{-0.4cm}

Interactions in the medical field are currently limited to direct physical interactions between a physician and a model, such as scribbles \cite{wong_scribbleprompt_2023} or clicks \cite{chen_focalclick_2022} that are typically performed using mouse movements or mouse clicks. 
A downside of this approach is that these methods cannot be used in situations where physicians need to use their hands to perform treatments or surgeries while depending on precise, problem-tailored segmentations. Typical examples in the clinical routine are orthopedic surgeries such as the insertion of implants~\cite{KUMAR2021216} that require intraoperative CT images, real-time imaging in endoscopy~\cite[p. 443--450]{dossel_bildgebende_2016} or real-time X-rays during cardiac catheterization~\cite{MCLAUGHLIN2006531}. To address the shortcomings of current physical interactive segmentation models, this work pioneers the development of a model that can work with natural language. In this work, we address the primary challenge of designing a system that effectively utilizes natural language for segmentation and interaction tasks. We make significant progress towards this goal by first developing a framework that works with text-based inputs, laying the groundwork for future adaptation to spoken language, which given the robust capabilities of existing Voice2Text models can expected to be seamless.

%TODO: Alex. Yolo ansprechen?

Within this work, we introduce LIMIS: A \textbf{L}anguage-based
\textbf{I}nteractive
\textbf{M}edical \textbf{I}mage \textbf{S}egmentation framework which allows users to generate an initial segmentation mask using natural language and perform interactive improvements upon potential errors using natural language. 
Our contributions are summarized as follows: (1) We develop a segmentation pipeline that is able to create an initial segmentation mask from natural language by adapting Grounded SAM~\cite{idea_groundedsam_2024} to the medical domain. 
(2) We pioneer language-based interaction, allowing the users to adapt the initial segmentation mask to incorporate their knowledge into the segmentation mask by using only language.
(3) We validate the segmentation performance of our approach across multiple medical datasets.
(4) We validate the suitability of LIMIS' interactive capabilities via a user study with professional radiologists. 

\vspace{-0.1cm}
\section{Methods}
\label{sec:methods}
\begin{figure*}
    \centering
    \includegraphics[trim=0 10 105 250, clip, width=0.9\textwidth]{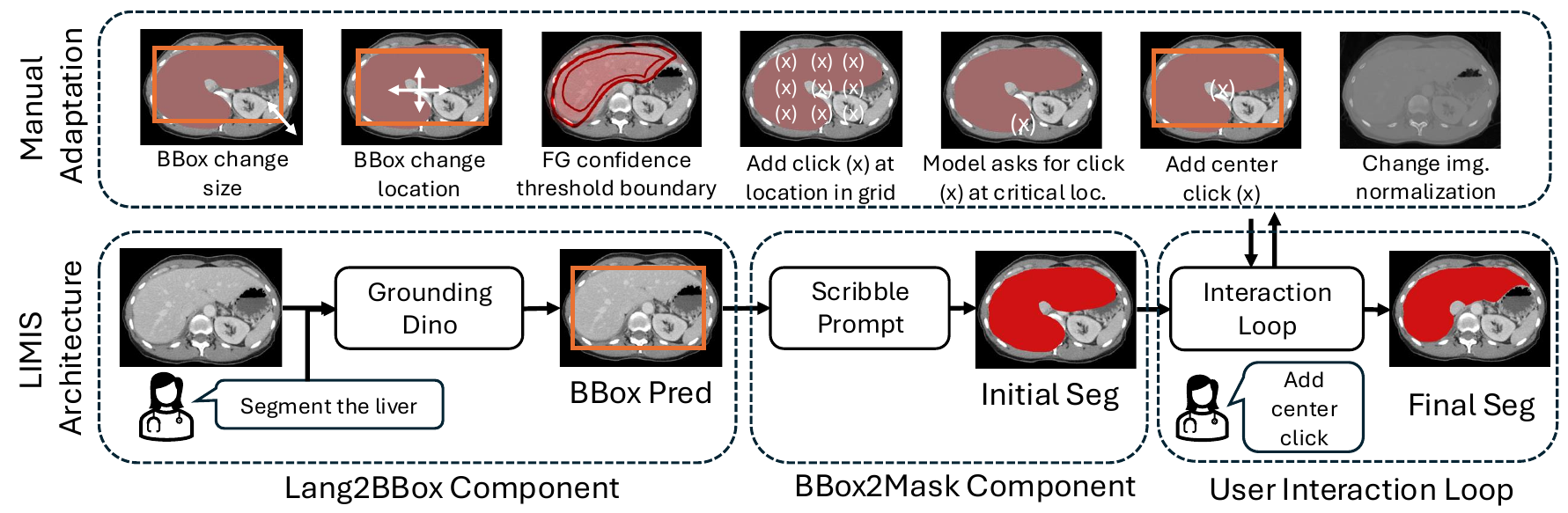}
        \caption{Top: Manual Language-based Adaptation options. Bottom:  LIMIS flowchart showing user input processing from language prompt to final mask via Grounding DINO (Lang2BBox), ScribblePrompt (BBox2Mask), and User Interaction Loop.
        }
        \vspace{-0.2cm}
    \label{fig:pipeline:structure}
\end{figure*}

This section introduces the proposed LIMIS architecture. It consists of three major components: the Language to Bounding Box component (Lang2BBox), which works with the Bounding Box to Segmentation component (BBox2Mask) to generate an initial segmentation, and the User Interaction Loop. 
Fig.~\ref{fig:pipeline:structure} shows the structure of the LIMIS architecture including some of the manual user interactions. 

\vspace{-0.1cm}
\subsection{Generating an Initial Segmentation from Language}
To generate an initial segmentation mask from language input, we draw inspiration from the Grounded SAM~\cite{idea_groundedsam_2024} architecture which has already been explored for colonoscopy~\cite{biswas_polyp-sam_2023} or X-Ray~\cite{kamaleswaran_lung_nodate}. Contrary to these works, we do not keep the standard Grounded SAM architecture but adapt both its components: SAM~\cite{kirillov_segment_2023} since it has been shown to perform poorly on non-optical medical images such as radiographic images~\cite{wong_scribbleprompt_2023}, and Grounding DINO~\cite{liu_grounding_2023}. %Besides the segmentation model, we make sure that the entire Grounded SAM architecture is tailored to the domain of CT (Computed Tomography) imaging by also adapting Grounding Dino, as we explain in the following section.

To obtain an initial segmentation, we first generate a bounding box from a language prompt in the Lang2BBox component. To achieve this, we adapt the text-based object detector Grounding DINO~\cite{liu_grounding_2023} to the medical domain using the parameter efficient fine-tuning method LoRA~\cite{hu_lora_2021}. This LIMIS component predicts a bounding box around the target object. In the BBox2Mask component, we use the predicted bounding box as a prompt to the ScribblePrompt \cite{wong_scribbleprompt_2023} model, which is a medical adaptation of SAM~\cite{kirillov_segment_2023}, to predict an initial segmentation mask.
%The segmentation pipeline consists of three parts: The object detector Grounding DINO \cite{liu_grounding_2023}, SAM, and user interactions.  Grounding DINO and SAM combined is called Grounded SAM and is used to predict an initial segmentation mask. Grounding DINO predicts a bounding box around the object and this bounding box is used as input for SAM that predicts the segmentation mask. A text prompt and an image are used as input for Grounded SAM. The user interactions improve upon the predicted initial segmentation mask. The text prompt is the name of an organ, e.g. liver.

%Instead of the original SAM, we use two versions adapted to medical images: MedSAM \cite{ma_segment_2023} and ScribblePrompt \cite{wong_scribbleprompt_2023}. We analyzed how the SAM version influences the result of the whole pipeline. The input images of MedSAM are interpolated to the required size of 1024$\times$1024 pixels with a bicubic interpolation. ScribblePrompt needs an input image size of 128$\times$128 pixels. This leads to a loss of details in the segmentation mask. Therefore, we use only a part of the image as input for ScribblePrompt. This part is created by cropping the image around the bounding box predicted by Grounding DINO. We choose a margin around the bounding box depending on the bounding box size with a minimum of 10 pixels added on each side of the bounding box.
\vspace{-0.1cm}
\subsection{Segmentation Refinement through User Interactions}
The third component of LIMIS is the User Interaction Loop, allowing refinements of the initial segmentation mask via user interactions. 
It starts by applying a default adaptation to the image and segmentation mask. Users then assess if this improves the segmentation mask and choose whether to keep it. This default strategy normalizes the CT image based on the target organ's typical radiological visualization parameters, e.g., using liver-specific CT window settings. The strategy further expands the bounding box by 10 pixels on each side. The choices for these default values are ablated in~\Cref{results:GroundedSAM}.

After applying the default options, users have two methods to address potential segmentation mask errors:
\begin{itemize}
    \item \textbf{Manual Adaptation}: Adjust the segmentation mask through manual interactions.
    \item \textbf{Automated Multi-Step Strategies}: Choose from four predefined automated strategies designed to correct common segmentation issues.
\end{itemize}

\noindent Throughout the segmentation process, users can decide after each interaction whether to continue with the updated mask or revert to any previous version. The final segmentation mask can be selected from any step, and it does not need to be the one generated in the last interaction.

\vspace{-0.1cm}
\subsubsection{Manual Adaptation via Interactions}

%Apart from the predefined suggestions, the users can choose from the manual interactions with a language prompt. 
%TODO: Lena und Alex entwurf für Interaction Figure
LIMIS offers manual interactions inspired by physical click-based interactions and active learning regimes:
\begin{itemize}
    \item \textbf{Bounding Box Changes}: Shift location or change size.
    \item \textbf{Confidence Threshold}: Change the threshold determining if critical pixels are part of the foreground mask.
    \item \textbf{Click in Grid}: Add a foreground/background click in one of 16 locations organized as a regular grid.
    \item \textbf{Critical Region Decision}: The system asks the users to decide for specific critical points if these belong to the foreground structure or the background.
    \item \textbf{Center Click}: Add a foreground click in the center of the bounding box.
    \item \textbf{Change Normalization}: Choose a new CT visualiztion window (location \& width) for image normalization.
    \item \textbf{Generate Examples}:
    Show exemplary interactions.
    \item \textbf{Remove Component}: Remove a connected component.
    \item \textbf{Ensemble}: Combine the segmentation masks of the following interactions: box size change, center click, and change of normalization.
\end{itemize}

%\noindent  We show an overview of the entire LIMIS model framework in Figure \ref{fig:pipeline:structure} including manual adaption options.
\vspace{-0.1cm}
\subsubsection{Problem-oriented, guided multi-step Interactions}
Besides the manual interactions, four problem-oriented, predefined multi-step adaptations guide the user as shown in table~\ref{fig:title-fig} on how to refine the initial segmentation mask:
\begin{itemize}
    \item \textbf{Wrong image part segmented}: Add center click, adjust normalization, and add grid points.
    \item \textbf{Target oversegmented}:
    Increase the foreground confidence threshold and add critical points and grid points.
    \item \textbf{Target Undersegmented}:
    Increase BBox, reduce foreground confidence threshold, and add critical points.
    \item \textbf{Target has low HU-values}: Adapt image normalization.%Adapt image normalization to lighten dark target area.
\end{itemize}
%that the segmented area is too large or too small, and that the target area is dark. Users choose from these four suggestions with a language prompt. 

\noindent In each of the four suggestions, the predefined manual interactions guide the users, thereby streamlining the segmentation process and helping the users to familiarize themselves with the effects of the manual interactions used during the automated processes. 

\vspace{-0.1cm}
\subsection{Adaptation Strategy Grounding DINO (Lang2BBox)}
Within the following section, we outline our proposed adaptation strategy of the non-medical Grounding DINO object detector to the medical domain. 
\label{chap:Training_GDINO}

\noindent \textbf{Changes to Network Structure}:
We use the SOTA parameter efficient fine-tuning approach LoRA~\cite{hu_lora_2021} to adapt Grounding DINO to the medical domain. Compared to other domain adaptation methods such as adapters, it does not add any additional inference time. We include LoRA layers to the self-attention and deformable self-attention layers within the Grounding DINO architecture.

\noindent \textbf{Data}:
We use three publicly available medical CT datasets for this work: DAP Atlas~\cite{jaus_towards_2023}, TotalSegmentator~\cite{wasserthal_totalsegmentator_2022} and WORD~\cite{luo_word_2022}. 
In this work, we only use the anatomical structures available in all three datasets: esophagus, stomach, duodenum, colon, gallbladder, liver, pancreas, kidney left, kidney right, bladder, and spleen. %The pooled dataset is split into 80\% for training, 10\% for validation, and 10\% for testing, ensuring no overlap in validation and test set with images seen by ScribblePrompt during its training.
Each dataset is initially split into 80\% training, 10\% validation, and 10\% testing. The resulting subsets are then pooled across all datasets, maintaining the same proportions. We make sure our test and validation sets have no overlap with images used by the authors during the ScribblePrompt model training

\noindent \textbf{Data Pre-Processing}:
Images are pre-processed by slicing CT volumes into 2D images along the transversal plane. Following nnUNet~\cite{isensee_nnu-net_2021}, we clip the HU-values to the 0.5 and 99.5 percentiles. We normalize using the mean and standard deviation of the foreground pixels. To address dataset differences, we commit to a common pixel spacing, image size, and image orientation. As data augmentations, we use image translations, rotations and scaling with an individual probability of 10\%. The range of rotation is -10.3$^\circ$ to 10.3$^\circ$, the translation up to 10 pixels and the scaling factor is between 0.9 and 1.1.

\noindent \textbf{Language Prompt Generation}:
The training of Grounding DINO requires a language input which we model as a sequence of label names that consists of two parts. The first part is the label names of the organs present in the image. We further add random label names from all training classes that are not present in the image, simulating noise in the language prompts. All label names in the prompt are shuffled randomly.

\noindent \textbf{Loss Function and Hyperparameters}.
The loss function and most hyperparameters are chosen according to \cite{liu_grounding_2023}. A detailed summary of the ablated training configuration is shown in~\cref{tab:results:confs_training}.

\vspace{-0.1cm}
\section{Experiments and Results}
\label{sec:results}
\vspace{-0.1cm}
\subsection{Grounding DINO: Implementation \& Evaluation}
The fine-tuning of Grounding DINO was conducted on three NVIDIA RTX 6000 GPUs with an individual batch size of 64 per GPU, yielding a total batch size of 192. The model achieved a mean Average Precision (mAP) of 0.54, with mAP@50 at 0.80 and mAP@75 at 0.58.

\noindent \textbf{Ablations}
We ablated the usage of augmentations (augm), the learning rate (lr), and the number of additional label names that were added to the text prompt (num add lab). Table \ref{tab:results:confs_training} shows the influence of these hyperparameters on the results of the training. Configuration 1 achieves the highest mAP. We find that applying augmentations generally leads to improved results, and using a greater number of random label names outperforms using fewer.
\begin{table}[h!]
    \centering
    \footnotesize
    \caption{Tested hyperparameter configurations on val set. %The hyperparameters are the usage of augmentations (augm), the learning rate (lr) and the number of random additional categories added to the text prompt (num add cat).
    }
    \centerline{
    \begin{tabular}{ccccc} \toprule
         \textbf{Config} & \textbf{augm} & \textbf{lr} & \textbf{num add lab} & \textbf{mAP} \\
         \midrule
         1& yes & 1e-4 &  8 & \textbf{0.541}\\
         2&yes & 1e-4 & 2 & 0.540\\
         3 & no & 1e-4 & 8 & 0.525 \\
         4 & no & 1e-4 & 2 & 0.510\\
         5 & yes & 1e-5 & 2 & 0.499\\ \bottomrule
    \end{tabular}}
    \label{tab:results:confs_training}
\end{table}
%\vspace{-0.5cm}

\subsection{ScribblePrompt: Implementation \& Evaluation}
\label{results:GroundedSAM}
We evaluate ScribblePromt~\cite{wong_scribbleprompt_2023} as our BBox2Seg component for different configurations. 
We compare feeding the entire image with its bounding box to the model as well as the image cropped to the bounding box plus a small margin around it with the latter setting leading to significantly higher Dice scores ($53\%$ vs. $58\%$) across all segmented organs. We further identify that using common radiologist CT visualization windows as the input to ScribblePrompt boosts performance from $58\%$ Dice to $63\%$. Finally, we investigate if the predicted bounding box should be enlarged by default by a small number of pixels. We find that on average increasing the bounding box by $10$ pixels on each side improves the performance to $66\%$ Dice. Enlarging the box further to $20$ pixels per side decreases the performance significantly to $54\%$ Dice indicating a worse localization cue by the enlarged bounding box. We show the effect of the stated default option qualitatively in~\cref{fig:results:stepsLiver} (default).

\vspace{-0.1cm}
\subsection{Interaction Loop: Evaluation via User Study}
The third component of the LIMIS architecture is the User Interaction Loop. We evaluate its performance via a user study with four participants: Two radiologists, one medical doctor, and one medical student. 
%The users had 10 minutes to familiarize themselves with the system and were then asked to segment as many images as possible with the best possible result in 50 minutes. The users should use a maximum of 5 minutes per image and move on to the next image if two consecutive interactions did not improve the segmentation mask. 
We present the users with a series of CT images from our test set in which they are tasked to segment one anatomical structure. We design the user interaction interface as a GUI facilitating using the system for non-technical users. During the user study, the participants collectively annotated 63 images. We evaluate the results of the user study and find that for 41 images (65\%), the final segmentation had a higher Dice score than the initial segmentation. The average Dice improvement for these images was $(6\pm5.13)\%$. Around 21\% of the images had a lower final Dice score $(-2\pm 2)\%$ and 14\% of the images resulted in identical Dice scores pre- and post-interactions. Overall, the Dice score change was $(4\pm 7.0)\%$. It however has to be pointed out that the participants were not forced to submit the mask after the last interaction but were allowed to submit any intermediate and even the initial prediction. Thus, some Dice score drops may reflect differing expert opinions, not system weaknesses. Additionally, it has to be acknowledged that the overall performance of LIMIS is limited by the ScribblePrompt foundation model used as the BBox2Mask component.

%thus some of the Dice score changes, in particular, the ones which are lower than the initial dice score may not reflect a weakness of the system but rather a different expert opinion on how the target segmentation should look. It also has to be acknowledged that the overall performance of the LIMIS system is capped by the performance of the ScribblePrompt foundation model serving as our BBox2Mask component.

\begin{figure}%[h]
\centering
    \begin{subfigure}{0.23\textwidth}
        \includegraphics[width=\textwidth]{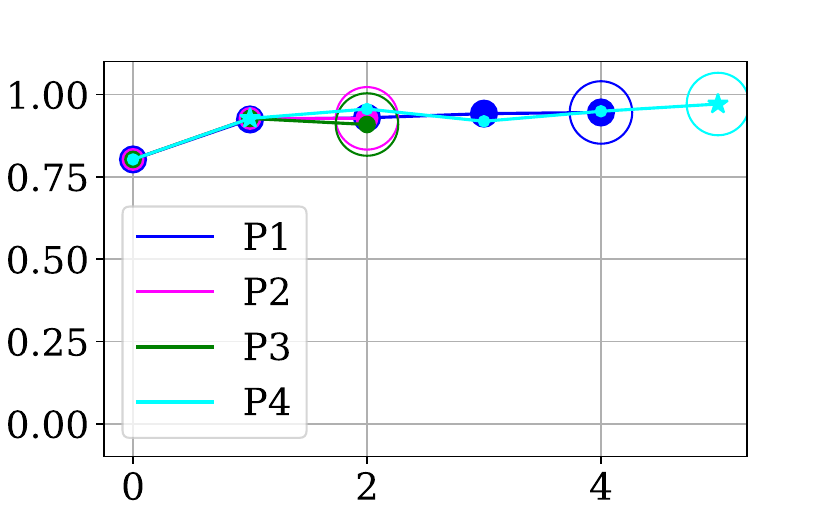}
        \caption{Atlas, bladder}
        \label{fig:results:mainstudy:DSCvsSteps:AtlasBladder}
    \end{subfigure}
    \hfil
    \begin{subfigure}{0.23\textwidth}
        \includegraphics[width=\textwidth]{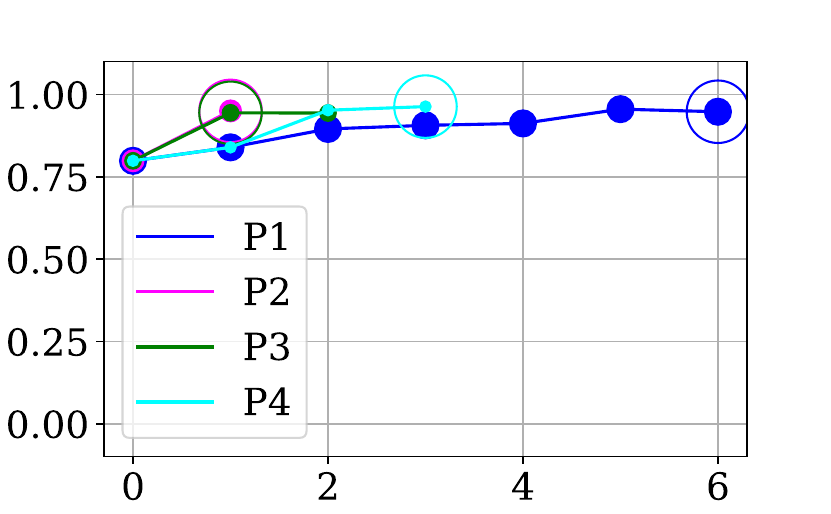}
        \caption{TotalSegmentator, liver}
        \label{fig:results:mainstudy:DSCvsSteps:TSLiver}
    \end{subfigure}
\caption[Dice score over interaction step.]{Dice score over interaction steps for two images. Step 0 is the initial mask; if ``default" was accepted, it's step 1.%Subtitles show the dataset and segmented category. 
Big circles mark the user’s final chosen mask. Stars indicate when a non-latest step was adapted, marking both the adapted and resulting steps.% Best viewed in color and with zoom.
}
    \label{fig:results:mainstudy:DSCvsSteps}
    \end{figure}

\begin{figure}%[h]
    \centering
    \includegraphics[width=1\linewidth]{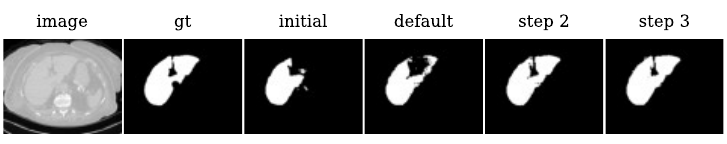}
    \caption{Liver segmentation mask over iteration steps. The first image shows the CT scan, the second the ground truth (gt), and the third the initial LIMIS prediction. ``Default" presents the mask after the default option, and the last two images show masks from steps 2 and 3.}
    \label{fig:results:stepsLiver}
    
\end{figure}

In \cref{fig:results:mainstudy:DSCvsSteps} we show the Dice scores change over the iteration steps when tasked to segment the bladder (left) within a sample taken from the DAP Atlas~\cite{jaus_towards_2023} dataset and the liver (right) from a TotalSegmentator~\cite{wasserthal_totalsegmentator_2022} sample. A qualitative example of the change of the segmentation mask is shown in~\cref{fig:results:stepsLiver}.

We evaluate the usability of LIMIS with the NASA TLX and the Single Ease Question (SEQ). Table \ref{tab:results:mainstudy:nasatlx} shows the participants' assessments of LIMIS.

The range of the participants' answers was wide for most of the questions.
P2, the most experienced radiologist with over 7 years of experience in annotating medical images, rated the system very favorable and liked the ``novelty of [the] segmentation approach with text". 
Although P1 rated LIMIS with high values for effort and frustration, the participant stated that ``once [...] [you] got into it, it was easy to use".
 Furthermore, the participants stated that the four predefined ``suggestions are very valuable".
\begin{table}[h]
    \centering
    \footnotesize
    \caption[Answers of the participants of the main study to the questionnaire.]{Participants' answers to NASA TLX and SEQ.}
    \begin{tabular}{lcccc} \toprule
          & \textbf{P1} & \textbf{P2} & \textbf{P3} & \textbf{P4}\\
         \midrule
         Mental Demand & 14 & 5& 11& 8 \\
         Physical Demand& 1 & 2 & 1& 4 \\
         Temporal Demand & 5 & 2 & 14& 5 \\
         Performance & 10 & 5 & 15& 10 \\
         Effort & 12 & 5 & 12& 10 \\
         Frustration & 14& 1 & 18& 10 \\
         \midrule
         SEQ & 4& 2& 5& 4 \\ \bottomrule
         
    \end{tabular}
   \vspace{-0.1cm}
    \label{tab:results:mainstudy:nasatlx}
\end{table}
%\vspace{-0.2cm}
\section{Discussion and Conclusion}
%The results show that the participants were able to use the interactions to improve upon the initial segmentation mask. There can be multiple reasons why there were final segmentation masks with lower Dice scores. The ground truth in medical image analysis is often ambiguous. Physicians disagree on how a segmentation mask should look like. Additionally, it was sometimes challenging to decide if an user interaction improved the segmentation mask if the changes were only small.

We present LIMIS, the first language-only interactive model for medical imaging. Adapting a Grounded SAM-inspired architecture, LIMIS integrates problem-oriented multi-step language interactions with state-of-the-art medical foundation models, enabling accurate initial segmentations and user-driven mask adaptations. LIMIS was tested on multiple datasets, and its usability was evaluated by medical experts.

\newpage

\noindent\textbf{Compliance With Ethical Standards:} This research study was conducted retrospectively using human subject data made available in open access. Ethical approval was not required as confirmed by the license attached with the open access data.
\noindent\textbf{Acknowledgments:} The present contribution is supported by the Helmholtz Association under the joint research school “HIDSS4Health – Helmholtz Information and Data Science School for Health. The user studies were done in collaboration with the
Annotation Lab Essen (annotationlab.ikim.nrw/)

\bibliographystyle{IEEEbib}
\bibliography{refs}

\begin{thebibliography}{10}

\bibitem{isensee_nnu-net_2021}
Fabian Isensee, Paul~F. Jaeger, Simon A.~A. Kohl, Jens Petersen, and Klaus~H. Maier-Hein,
\newblock ``{nnU}-{Net}: a self-configuring method for deep learning-based biomedical image segmentation,''
\newblock {\em Nature Methods}, vol. 18, no. 2, pp. 203--211, Feb. 2021.

\bibitem{wasserthal_totalsegmentator_2022}
Jakob Wasserthal, Hanns-Christian Breit, Manfred~T. Meyer, Maurice Pradella, Daniel Hinck, Alexander~Walter Sauter, Tobias Heye, Daniel Boll, Joshy Cyriac, Shan Yang, Michael Bach, and Martin Segeroth,
\newblock ``{TotalSegmentator}: Robust segmentation of 104 anatomical structures in {CT} images.,''
\newblock {\em Radiology. Artificial intelligence}, vol. 5(5), pp. e230024, 2022.

\bibitem{kirillov_segment_2023}
Alexander Kirillov, Eric Mintun, Nikhila Ravi, Hanzi Mao, Chloe Rolland, Laura Gustafson, Tete Xiao, Spencer Whitehead, Alexander~C. Berg, Wan-Yen Lo, Piotr Doll{\'a}r, and Ross~B. Girshick,
\newblock ``Segment anything,''
\newblock {\em 2023 IEEE/CVF International Conference on Computer Vision (ICCV)}, pp. 3992--4003, 2023.

\bibitem{wong_scribbleprompt_2023}
Hallee~E. Wong, Marianne Rakic, John Guttag, and Adrian~V. Dalca,
\newblock ``Scribbleprompt: Fast and flexible interactive segmentation for any biomedical image,''
\newblock {\em European Conference on Computer Vision (ECCV)}, 2024.

\bibitem{idea_groundedsam_2024}
Tianhe Ren, Shilong Liu, Ailing Zeng, Jing Lin, kunchang Li, He~Cao, Jiayu Chen, Xinyu Huang, Yukang Chen, Feng Yan, Zhaoyang Zeng, Hao Zhang, Feng Li, Jie Yang, Hongyang Li, Qing Jiang, and Lei Zhang,
\newblock ``{Grounded SAM}: Assembling open-world models for diverse visual tasks,''
\newblock {\em arXiv preprint arXiv::2401.14159}, 2024.

\bibitem{chen_focalclick_2022}
X.~Chen, Z.~Zhao, Y.~Zhang, M.~Duan, D.~Qi, and H.~Zhao,
\newblock ``Focal{C}lick: Towards practical interactive image segmentation,''
\newblock in {\em 2022 IEEE/CVF Conference on Computer Vision and Pattern Recognition (CVPR)}, Los Alamitos, CA, USA, 6 2022, pp. 1290--1299, IEEE Computer Society.

\bibitem{KUMAR2021216}
Vishal Kumar, Vishnu Baburaj, Sandeep Patel, Siddhartha Sharma, and Raju Vaishya,
\newblock ``Does the use of intraoperative {CT} scan improve outcomes in orthopaedic surgery? {A} systematic review and meta-analysis of 871 cases,''
\newblock {\em Journal of Clinical Orthopaedics and Trauma}, vol. 18, pp. 216--223, 2021.

\bibitem{dossel_bildgebende_2016}
Olaf Dössel,
\newblock {\em Bildgebende {Verfahren} in der {Medizin}},
\newblock Springer, Berlin, Heidelberg, 2016.

\bibitem{MCLAUGHLIN2006531}
Peter McLaughlin, Lee Benson, and Eric Horlick,
\newblock ``The role of cardiac catheterization in adult congenital heart disease,''
\newblock {\em Cardiology Clinics}, vol. 24, no. 4, pp. 531--556, 2006.

\bibitem{biswas_polyp-sam_2023}
Risab Biswas,
\newblock ``Polyp-{SAM}++: {Can} a text guided {SAM} perform better for polyp segmentation?,''
\newblock {\em arXiv preprint arXiv::2308.06623}, Aug. 2023.

\bibitem{kamaleswaran_lung_nodate}
Rishikesan Kamaleswaran, Pulakesh Upadhyaya, Rishika Iytha~Sridhar, and Dhanush~Babu Ramesh,
\newblock ``{Lung Grounded-SAM (LuGSAM)}: {A} novel framework for integrating text prompts to segment anything model {(SAM)} for segmentation task of {ICU} chest {X}-rays,''
\newblock {\em techrxiv preprint techrxiv.24224761.v1}.

\bibitem{liu_grounding_2023}
Shilong Liu, Zhaoyang Zeng, Tianhe Ren, Feng Li, Hao Zhang, Jie Yang, Chunyuan Li, Jianwei Yang, Hang Su, Jun Zhu, et~al.,
\newblock ``Grounding {DINO}: Marrying {DINO} with grounded pre-training for open-set object detection,''
\newblock {\em arXiv preprint arXiv:2303.05499}, 2023.

\bibitem{hu_lora_2021}
Edward~J. Hu, Yelong Shen, Phillip Wallis, Zeyuan Allen-Zhu, Yuanzhi Li, Shean Wang, Lu~Wang, and Weizhu Chen,
\newblock ``{LoRA}: Low-rank adaptation of large language models,''
\newblock {\em arXiv preprint arXiv::2106.09685}, June 2021.

\bibitem{jaus_towards_2023}
Alexander Jaus, Constantin Seibold, Kelsey Hermann, Alexandra Walter, Kristina Giske, Johannes Haubold, Jens Kleesiek, and Rainer Stiefelhagen,
\newblock ``Towards unifying anatomy segmentation: {Automated} generation of a full-body {CT} dataset via knowledge aggregation and anatomical guidelines,''
\newblock {\em arXiv preprint arXiv::2307.13375}, July 2023.

\bibitem{luo_word_2022}
Xiangde Luo, Wenjun Liao, Jianghong Xiao, Jieneng Chen, Tao Song, Xiaofan Zhang, Kang Li, Dimitris~N. Metaxas, Guotai Wang, and Shaoting Zhang,
\newblock ``{WORD}: {A} large scale dataset, benchmark and clinical applicable study for abdominal organ segmentation from {CT} image,''
\newblock {\em Medical Image Analysis}, vol. 82, pp. 102642, Nov. 2022.

\end{thebibliography}

\end{document}